\documentclass[10pt, conference, letterpaper]{IEEEtran}
\IEEEoverridecommandlockouts

\usepackage[letterpaper, top=0.75in, bottom=1.05in, left=0.625in, right=0.625in]{geometry}
\usepackage{cite}
\usepackage{amsmath,amssymb,amsfonts}
\usepackage{algorithmic}
\usepackage{graphicx}
\usepackage{textcomp}
\usepackage{comment}
\usepackage{xcolor}
\usepackage{subcaption}
\usepackage{booktabs}
\def\BibTeX{{\rm B\kern-.05em{\sc i\kern-.025em b}\kern-.08em
    T\kern-.1667em\lower.7ex\hbox{E}\kern-.125emX}}
\usepackage[draft, colorlinks,linkcolor = black,anchorcolor = black,citecolor = black]{hyperref}

\begin{document}

\title{PRI-Net: A Lightweight Multimodal Framework\\for 3D UAV Localization
}
\author{
    \IEEEauthorblockN{Zhixuan Chen$^{*}$, Jialiang Lu$^{*}$, Zhong Ye$^{\dagger}$, Yinghui He, Guanding Yu,~\IEEEmembership{Senior Member,~IEEE}}
    \IEEEauthorblockA{College of Information Science and Electronic Engineering, Zhejiang University, Hangzhou, China\\
Zhejiang Key Laboratory of Multimodal Communication Networks and Intelligent Information Processing, Hangzhou, China}
    \IEEEauthorblockA{Email: \{3230104529,3230104505,12431118,2014hyh,yuguanding\}@zju.edu.cn, $^{*}$Equal contribution, $\dagger$~Corresponding author} 
}

\maketitle

\begin{abstract}
Accurate 3D localization of unmanned aerial vehicles (UAVs) remains challenging for existing multimodal approaches due to sparse LiDAR geometry, modality-imbalanced fusion, and redundant feature transmission over constrained edge-to-server links. To address these limitations, we propose PRI-Net, an efficient and lightweight multimodal fusion framework for UAV localization that integrates point cloud splatting, residual attention fusion, and an information bottleneck. Specifically, a 3D point cloud splatting (3DPCS) strategy is introduced to transform sparse LiDAR observations into geometrically consistent dense depth maps. A residual attention fusion (RAF) module is then designed to alleviate modal bias by using an image branch for coarse estimation and a gated fusion branch for refinement. In addition, a multimodal information bottleneck (MIB) module compacts features by filtering task-irrelevant redundancy. Experiments show that PRI-Net achieves high localization accuracy with lightweight architectures, while reducing feature dimensionality and improving edge-to-server UAV sensing efficiency and robustness.
\end{abstract}


\section{Introduction}
Low-altitude unmanned aerial vehicle (UAV) sensing systems have developed rapidly, making accurate UAV localization essential for monitoring, cooperative perception, and situational awareness. Among ground-based modalities, vision-based methods are widely studied for their low cost and mature pipelines~\cite{he2025task}. YOLO-based approaches have achieved accurate real-time UAV detection and tracking~\cite{8999675,bao2024uav}, while enhanced vision networks address small targets, scale variation, and complex illumination~\cite{zhang2023drone}. However, image-based methods lack explicit depth and rely on appearance cues, limiting robustness in challenging environments. Therefore, LiDAR-based methods have been introduced to exploit 3D spatial structure. PointNet~\cite{qi2017pointnet}, PointNet++~\cite{qi2017pointnetplusplus}, and PointRCNN~\cite{zhou2022point} have advanced point cloud representation and 3D object detection. Despite their geometric advantages, LiDAR-based methods often face sparse observations, limited semantic richness, and high complexity, hindering deployment on resource-constrained edge devices. Recent graph-based localization systems have also improved robustness to dynamic sensing configurations and cross-scenario variations~\cite{11278769,11268513}. However, they mainly focus on CSI-based wireless localization rather than vision-LiDAR UAV perception. Therefore, multimodal fusion of images and point clouds has emerged as a promising direction for exploiting their complementary strengths.

Multimodal fusion methods have therefore attracted increasing attention. MV3D~\cite{chen2017multi} proposed an multi-view fusion framework, PointPainting~\cite{vora2020pointpainting} introduced a lightweight semantic enhancement strategy, and BEVFusion~\cite{liu2023bevfusion} developed a unified bird's-eye-view representation. Although these methods perform well, several challenges remain. First, when point clouds are projected onto the image plane, their inherent sparsity often produces depth maps with severe holes, limiting geometric representation. Second, large distribution gaps across modalities can lead to inconsistent feature representations and thus reduce fusion effectiveness. Third, existing multimodal methods often rely on heavy feature extraction and fusion pipelines, increasing computation overhead and imposing substantial transmission costs in edge-to-server deployment scenarios.

To address these challenges, we propose PRI-Net, named after its three key components: point cloud splatting, residual attention fusion, and information bottleneck. Specifically, to overcome depth incompleteness caused by point cloud sparsity, we introduce a 3D point cloud splatting (3DPCS) method that densifies projected depth maps by updating neighboring pixels within a fixed-radius region. A residual attention fusion (RAF) module is designed to alleviate feature discrepancy and fusion imbalance through gated cross-modal interaction, where the image modality provides the primary estimate and the depth modality refines it through residual correction. Moreover, a multimodal information bottleneck (MIB) module is incorporated before transmission to reduce communication overhead in edge-cloud scenarios by compressing features while preserving task-relevant information~\cite{tishby2000information,alemi2016deep,mai2022multimodal}. The main contributions of this paper are summarized as follows:

\begin{figure*}[!t]
\centering
\includegraphics[width=0.8\textwidth]{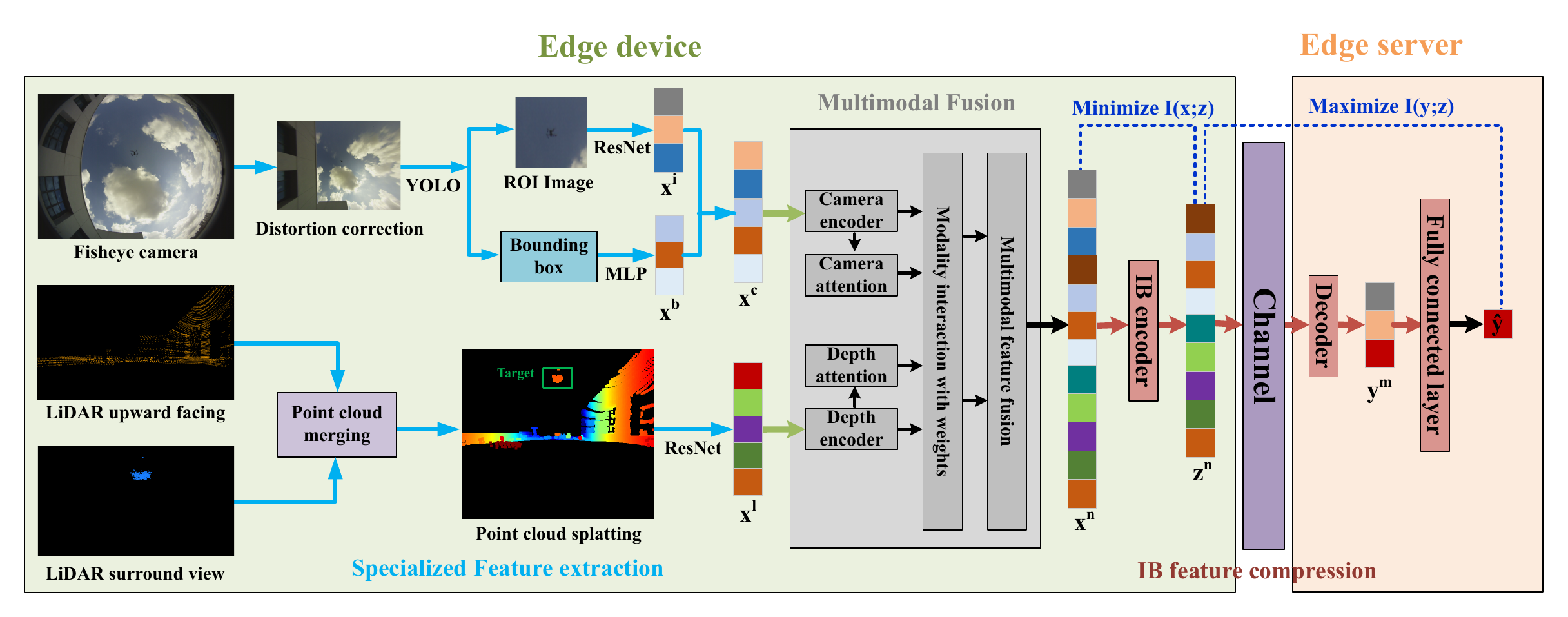}
\vspace{-2ex}
\caption{Framework of the proposed PRI-Net.}
\label{fig_framework}
\vspace{-3ex}
\end{figure*}

\begin{itemize}

\item We propose the 3DPCS method to densify projected depth maps, alleviating projection holes and improving geometric representation.

\item We propose RAF module that uses the image branch for primary prediction and the depth branch for residual correction, enabling effective cross-modal complementarity.

\item We introduce the MIB module to compress transmitted features, reducing communication overhead while preserving task-relevant information.

\end{itemize}

The remainder of this paper is organized as follows. The system architecture is described in Section \ref{architecture}. The system model is discussed in Section \ref{model}. The proposed methods are explained in Section \ref{method}. Experiment results are reported in Section \ref{result} and the whole paper is concluded in Section \ref{conclusion}.

\section{System Architecture}
\label{architecture}


The proposed PRI-Net framework enables accurate ground-based UAV localization through an edge-to-server pipeline that processes high-bandwidth camera and LiDAR data. To reduce latency, computationally intensive feature extraction is performed at the edge device, while the edge server conducts final coordinate regression using compressed task-relevant representations.

As shown in Fig.~\ref{fig_framework}, the edge device integrates multimodal processing, adaptive fusion, and information-theoretic compression. First, a frozen detector identifies the region of interest (ROI) in rectified visual streams, while a splatting strategy converts sparse LiDAR points into dense depth maps. Together with the bounding box geometry, these inputs are fed into a multi-branch encoder to project visual semantics and absolute scale information into a latent space.

The RAF module is then employed to model cross-modal interactions through a gated mechanism, followed by a two-stage residual regression process in which visual estimates are refined by geometric cues. To further reduce transmission overhead, a compression module removes redundant information via the variational information bottleneck (VIB) ~\cite{alemi2016deep} framework. This partitioned architecture allows the server to receive only essential task-relevant features, thereby balancing localization accuracy and real-time communication efficiency.

\section{System Model}
\label{model}

We consider a multimodal perception system consisting of a camera and a LiDAR, with spatial calibration and temporal synchronization to ensure cross-modal alignment. To establish the geometric foundation for fusion, we first define the projection and ranging models of the two sensors. Rigid-body transformations are then introduced for coordinate alignment, so that heterogeneous observations can be represented in a unified form.

\subsection{Camera Subsystem}

The camera subsystem captures visual semantics by projecting 3D points onto a 2D pixel grid through sequential transformations from the camera coordinate system to the pixel coordinate system. 
We define the camera coordinate system as $\mathcal{C}$, where a point is represented as $\mathbf{p}^{\mathrm{c}} = [x^{\mathrm{c}}, y^{\mathrm{c}}, z^{\mathrm{c}}]^\top\in\mathbb{R}^3$.

To accommodate various lens types, we employ Mei’s unified projection model \cite{mei2007single}, which is particularly suited for UAV platforms where wide-angle lenses are frequently utilized to ensure a sufficient field of view across expansive flight envelopes. 
In the first stage of the transformation, the 3D point $\mathbf{p}^{\mathrm{c}}$ is mapped to the normalized image coordinates $\tilde{\mathbf{m}}$ via
\begin{equation}
\tilde{\mathbf{m}} = \left[ \dfrac{x^{\mathrm{s}}}{z^{\mathrm{s}} + \xi}, \dfrac{y^{\mathrm{s}}}{z^{\mathrm{s}} + \xi}, 1 \right]^\top =  \left[ x^{\mathrm{m}}, y^{\mathrm{m}}, 1 \right]^\top,
\end{equation}
where $\left[ x^{\mathrm{s}}, y^{\mathrm{s}}, z^{\mathrm{s}}\right]=\mathbf{p}^{\mathrm{c}}/\left\Vert\mathbf{p}^{\mathrm{c}}\right\Vert$, $\xi$ is the parameter representing the lens geometry, and $\tilde{\mathbf{m}}$ denotes the homogeneous representation of $\mathbf{m}=\left[ x^{\mathrm{m}}, y^{\mathrm{m}}\right]^\top$.

Since the projection is subject to non-linear optical artifacts, we apply a distortion model $\mathbf{m}^{\mathrm{d}} = D(\mathbf{m})$, which is explicitly defined as
\begin{equation}
\begin{cases} x^{\mathrm{d}} = x^{\mathrm{m}}(1 + k_1\rho^2 + k_2\rho^4) + 2p_1x^{\mathrm{m}}y^{\mathrm{m}} + p_2(\rho^2 + 2(x^{\mathrm{m}})^2) \\ y^{\mathrm{d}} = y^{\mathrm{m}}(1 + k_1\rho^2 + k_2\rho^4) + p_1(\rho^2 + 2(y^{\mathrm{m}})^2) + 2p_2x^{\mathrm{m}}y^{\mathrm{m}} ,\end{cases}
\end{equation}
where $\mathbf{m}^{\mathrm{d}} = [x^{\mathrm{d}}, y^{\mathrm{d}}]^\top$ denotes the distorted coordinates on the normalized image plane, $\rho^2 = (x^{\mathrm{m}})^2 + (y^{\mathrm{m}})^2$, and the coefficients $k_i, p_i \;(i=1,2)$ represent the radial and tangential distortion components, respectively. Finally, the mapping to the pixel coordinates $\tilde{\mathbf{u}} = [u, v, 1]^\top$ is achieved through the intrinsic mapping
\vspace{-2mm}
\begin{equation}
\tilde{\mathbf{u}} = \mathbf{K} \begin{bmatrix} x^{\mathrm{d}} \\ y^{\mathrm{d}} \\ 1 \end{bmatrix}, \quad \mathbf{K} = \begin{bmatrix} f^{\mathrm{x}} & 0 & c^{\mathrm{x}} \\ 0 & f^{\mathrm{y}} & c^{\mathrm{y}} \\ 0 & 0 & 1 \end{bmatrix},
\end{equation}
where $f^{\mathrm{x}}$ and $f^{\mathrm{y}}$ denote the focal lengths (in pixels) along the \pagebreak $u$ and $v$ axes, respectively, while $c^{\mathrm{x}}$ and $c^{\mathrm{y}}$ represent the coordinates of the principal point.

\subsection{LiDAR Subsystem}

We define the LiDAR coordinate system as $\mathcal{L}$, where each captured point $\mathbf{p}^{\mathrm{l}} = [x^{\mathrm{l}}, y^{\mathrm{l}}, z^{\mathrm{l}}]^\top \in \mathbb{R}^3$ is derived from the range $r$ integrated with the horizontal azimuth $\theta$ and vertical elevation $\phi$ of the laser beam. The transformation to the Cartesian representation in the LiDAR coordinate system is expressed as
\begin{equation}
\mathbf{p}^{\mathrm{l}} = \begin{bmatrix} x^{\mathrm{l}} \\ y^{\mathrm{l}} \\ z^{\mathrm{l}} \end{bmatrix} = \begin{bmatrix} r \cos \phi \cos \theta \\ r \cos \phi \sin \theta \\ r \sin \phi \end{bmatrix},
\end{equation}
where the range $r$ is determined by the time-of-flight principle, calculated as
\begin{equation}
r = \frac{c \cdot \Delta t}{2},
\end{equation}
where $c$ denotes the speed of light and $\Delta t$ represents the pulse travel time. Unlike the measured range, the angles $\theta$ and $\phi$ are determined by the LiDAR’s scanning mechanism. Owing to discrete angular sampling, the resulting point cloud is inherently unstructured and non-uniform, and its spatial sampling density decreases quadratically with the range $r$.

\subsection{Coordinate Transformation and Data Representation}

For practical UAV applications, observations must be represented in a unified world coordinate system $\mathcal{W}$. To this end, rigid-body transformations are defined between the world, LiDAR, and camera coordinate systems, so that points from different sensors can be mapped into a common coordinate system. This explicit geometric alignment simplifies cross-modal fusion by avoiding implicit coordinate learning in the network.

Assuming ideal temporal synchronization between the two sensors, the system performs simultaneous sampling to generate a sequence of $N$ multimodal pairs $(\mathbf{I}_{i}, \mathbf{P}_{i})$, where $i$ denotes the discrete time index. For each synchronized sample, the inputs are defined as follows

\begin{itemize}

\item \textbf{Image Input}: The image is represented as a 3D tensor $\mathbf{I}_{i} \in \mathbb{R}^{C \times H \times W}$, where $H \times W$ denotes the resolution and $C$ represents the number of channel dimensions. 

\item \textbf{Point Cloud Input}: The point cloud is represented as a sparse set $\mathcal{P}_i = \{ \mathbf{p}^{\mathrm{l}}_{i,j} \}_{j=1}^{N_i}$ in the LiDAR coordinate system $\mathcal{L}$, where $j$ denotes the point index within the set. For efficient computation, it is organized as a matrix $\mathbf{P}_i \in \mathbb{R}^{N_i \times 3}$, where each row corresponds to a 3D point.

\end{itemize}

\section{Proposed Methods}
\label{method}


This section presents the proposed PRI-Net framework in three parts. Section \ref{MDPFE} introduces image ROI extraction and point cloud splatting for data preprocessing. Section \ref{RAFM} describes the RAF and regression strategy. Section \ref{IB} presents an information bottleneck module for compact task-relevant representations and efficient communication. 

\subsection{Multimodal Data Processing and Feature Extraction} \label{MDPFE}

\subsubsection{Image Modality}

The raw image $\mathbf{I}_{i}$ is first rectified to $\mathbf{I}^{\text{rec}}_{i}$ using the inverse intrinsics and distortion model $D(\cdot)$. A frozen YOLOv7 detector is then used to localize the UAV and crop the target region based on the bounding box
\vspace{-1pt}
\begin{equation}
\mathbf{b}_{i} = [u^{\mathrm{t}}_{i}, v^{\mathrm{t}}_{i}, u^{\mathrm{b}}_{i}, v^{\mathrm{b}}_{i}, u^\mathrm{c}_i, v^\mathrm{c}_i, w^\mathrm{b}_i, h^\mathrm{b}_i]^\top \in\mathbb{R}^8,
\end{equation}
\vspace{-1pt}
where $(u^{\mathrm{t}}_i, v^{\mathrm{t}}_i)$ and $(u^{\mathrm{b}}_i, v^{\mathrm{b}}_i)$ are corner coordinates, $(u^\mathrm{c}_i, v^\mathrm{c}_i)$ is the center, and $(w^\mathrm{b}_i, h^\mathrm{b}_i)$ denotes the dimensions. The corresponding sub-matrix is then extracted and rescaled to a cropped image $\mathbf{I}_{i}^{\text{crop}} \in \mathbb{R}^{C \times H^{\mathrm{c}} \times W^{\mathrm{c}}}$ with predefined height $H^{\mathrm{c}}$ and width $W^{\mathrm{c}}$.
The final image representation $\mathbf{x}^{\mathrm{c}}_i$ is formed by combining ResNet features from $\mathbf{I}_{i}^{\text{crop}}$ with multilayer perceptron (MLP)-encoded spatial priors from $\mathbf{b}_i$, integrating localized appearance and geometric context.

\subsubsection{LiDAR Modality}

Due to the extreme sparsity of UAV LiDAR data, conventional point-based architectures often fail to capture local geometry. We therefore propose the 3DPCS method that converts discrete points into a continuous representation. Specifically, a virtual pinhole camera with intrinsics $\mathbf{K}^\mathrm{v}$ and pose $[\mathbf{R}^\mathrm{v} | \mathbf{t}^\mathrm{v}]$ is defined, and each point $\mathbf{p}^\mathrm{l}$ is projected onto the image plane as
\vspace{-1pt}
\begin{equation}
\tilde{\mathbf{u}}^{\mathrm{v}}=\begin{bmatrix} u^{\mathrm{v}} \\ v^{\mathrm{v}} \\ 1 \end{bmatrix} = \dfrac{1}{z^\mathrm{c} }\mathbf{K}^\mathrm{v} \left( \mathbf{R}^\mathrm{v} \mathbf{p}^\mathrm{l} + \mathbf{t}^\mathrm{v} \right).
\end{equation}
\vspace{-1pt}
where $\tilde{\mathbf{u}}^{\mathrm{v}}$ denotes the homogeneous coordinates on the virtual image plane and $z^\mathrm{c}$ is the depth. 


To fill spatial gaps, a splatting mechanism assigns the depth $z^\mathrm{c}$ to pixels within a radius $R$. For a point set $\mathbf{P}_{i}$, the depth map $\mathbf{M}_{i}$ at pixel $\mathbf{u}^\mathrm{v}$ is defined as
\vspace{-1pt}
\begin{equation}
\mathbf{M}_{i}(\mathbf{u}^\mathrm{v}) = \min_{\mathbf{u}} \{ z^\mathrm{c}(\mathbf{u})\} \quad \text{s.t.}\quad  \text{dist}(\mathbf{u}^\mathrm{v}, \mathbf{u}) \le R ,
\end{equation}
\vspace{-1pt}
where $\text{dist}(\cdot)$ denotes the $L_\infty$ distance. To fully exploit the geometric and spatial contextual information, we construct a geometric tensor $\mathbf{M}_{i}^{\text{geo}} \in \mathbb{R}^{3 \times H \times W}$ by concatenating the normalized depth map $\mathbf{M}_{i}$ with the normalized 2D coordinate grid $(u', v')$. 
A typical AI network~\cite{jin2026profiles}, i.e., convolutional neural network (CNN) encoder with a specialized input stem and a pretrained ResNet backbone, is used to extract geometric features $\mathbf{x}^\mathrm{l}_i$. 
This design provides structural and scale information while reducing the representation gap between image and depth modalities. 

\vspace{-1mm}
\subsection{Residual Attention Fusion Module}
\label{RAFM}
\subsubsection{Cross-modal Attention Fusion}

The feature extraction networks map multimodal inputs into a shared latent space, where $\mathbf{x}^\mathrm{c}, \mathbf{x}^\mathrm{l} \in \mathbb{R}^{d}$ denote the image and depth features. We employ a gated attention-based fusion method by concatenating them as $\mathbf{x}^\mathrm{f} = [\mathbf{x}^\mathrm{c} ; \mathbf{x}^\mathrm{l}]$. A lightweight attention network then generates modality-specific gating weights
\begin{equation}
\mathbf{w}^{m}=\sigma\left(h_{\mathrm{mlp}}^{m}(\mathbf{x}^{\mathrm{f}};\theta_{\mathrm{att}}^{m})\right)\in\mathbb{R}^{d}, \quad m\in\{\mathrm{c},\mathrm{l}\},
\end{equation}
where $\sigma(\cdot)$ is the Sigmoid activation function and $h_{\mathrm{mlp}}^{m}(\cdot)$ is an MLP parameterized with parameters $\theta_{\mathrm{att}}^{m}$. 
To preserve the original modality information, a residual gating mechanism is used to update the features
\begin{equation}
\tilde{\mathbf{x}}^{m} = \mathbf{x}^{m} + \mathbf{x}^{m}\odot \mathbf{w}^{m},\quad m\in\{\mathrm{c},\mathrm{l}\},
\end{equation}
where $\odot$ denotes element-wise multiplication. The fused representation is then formed as $\mathbf{x}^{\mathrm{n}}=[\tilde{\mathbf{x}}^{\mathrm{c}};\tilde{\mathbf{x}}^{\mathrm{l}}] \in \mathbb{R}^{2d}$.

\subsubsection{Residual Multimodal Regression}



Since image features are generally more stable, they may dominate training and suppress the depth modality. We therefore adopt a residual multimodal regression strategy. Specifically, the image branch first produces a base prediction $\hat{\mathbf{y}}^{\mathrm{c}}$, and the fusion branch then learns a residual correction $\Delta \mathbf{y}$, yielding
\begin{equation}
\hat{\mathbf{y}} = \hat{\mathbf{y}}^{\mathrm{c}} + \Delta \mathbf{y} = H_{\mathrm{c}}(\mathbf{x}^{\mathrm{c}}) + H_{\mathrm{F}}(\mathbf{x}^{\mathrm{n}}),
\end{equation}
where $H_{\mathrm{c}}(\cdot)$ and $H_{\mathrm{F}}(\cdot)$ denote the image regression network and the fusion-based residual network, respectively. This design allows the depth modality to refine the image prediction with complementary geometric information.

\subsubsection{Auxiliary Supervision}

To stabilize multimodal training, auxiliary supervision is further introduced for the fused, image, and depth branches. The overall training loss is defined as
\begin{equation}
L = \lambda_1 L_{\text{fusion}} + \lambda_2 L_{\text{img}} + \lambda_3 L_{\text{depth}},
\end{equation}
where $L_{\text{fusion}}$, $L_{\text{img}}$, and $L_{\text{depth}}$ represent the localization losses of the fused, image, and depth branches, respectively, with $\lambda_1, \lambda_2, \text{ and } \lambda_3$ serving as the corresponding weighting coefficients. 
This joint optimization strategy, combined with the residual regression mechanism, effectively mitigates modal imbalance and ensures stable multimodal predictions.

\subsection{Information Bottleneck Compression}
\label{IB}



The IB principle learns a compact representation that preserves task-relevant information while removing redundancy, making it suitable for UAV localization under limited computation and communication resources. To reduce the transmission overhead of high-dimensional fused features, we adopt an early-fusion multimodal information bottleneck (EMIB) \cite{mai2022multimodal} framework.

Specifically, the fused feature $\mathbf{x}^\mathrm{n}$ is fed into the IB module, with the localization label denoted by $\mathbf{y} \in \mathbb{R}^3$. Let $X$, $Y$, and $Z$ denote the random variables corresponding to $\mathbf{x}^\mathrm{n}$, $\mathbf{y}$, and the compressed latent representation $\mathbf{z}^\mathrm{n} \in \mathbb{R}^{k}$, respectively, where $k$ controls the bottleneck capacity. The IB objective is
\begin{equation}
\mathcal{L}_{\text{IB}} = I(Z;X) - \beta I(Z;Y),
\end{equation}
where $I(\cdot;\cdot)$ denotes mutual information and $\beta$ balances information compression and task relevance.

Since mutual information is generally intractable, VIB is adopted to obtain a tractable objective
\begin{equation}
\begin{aligned}
\mathcal{L}_{\mathrm{VIB}}
= {} & \mathbb{E}_{(\mathbf{x}^\mathrm{n},\,\mathbf{y})}
\mathbb{E}_{\mathbf{z}^\mathrm{n}\sim p(\mathbf{z}^\mathrm{n}\mid\mathbf{x}^\mathrm{n})}
\bigl[-\log q(\mathbf{y}\mid\mathbf{z}^\mathrm{n})\bigr] \\
& + \beta\, D_{\mathrm{KL}}\!\left(
p(\mathbf{z}^\mathrm{n}\mid\mathbf{x}^\mathrm{n})
\,\|\, r(\mathbf{z}^\mathrm{n})
\right).
\end{aligned}
\end{equation}
Where the first term corresponds to the task prediction loss and the second term enforces information compression.

In implementation, the posterior $p(\mathbf{z}^\mathrm{n}|\mathbf{x}^\mathrm{n})$ is parameterized by the network as a diagonal Gaussian distribution $\mathcal{N}(\boldsymbol{\mu}, \boldsymbol{\sigma}^2)$, where $\mu_j$ and $\sigma_j$ denote the mean and standard deviation of the $j$-th latent dimension, respectively. The KL divergence regularizes this posterior toward a prior distribution.The final training objective is given by
\begin{equation}
\mathcal{L}_{\text{ALL}} =
L_{\text{fusion}} +
\frac{\beta}{2}
\sum_{j=1}^{k}
\left(
\mu_j^2 + \sigma_j^2 - \log \sigma_j^2 - 1
\right).
\end{equation}

By introducing the IB constraint, the proposed framework learns a compact latent representation that preserves task-relevant information while suppressing redundant features. This mechanism effectively reduces communication overhead and encourages the model to focus on informative representations for UAV localization.

\section{Experimental Results}
\label{result}

This section evaluates the proposed PRI-Net framework through three ablation-style experiments: (i) analysis of input processing strategies and comparison between single-modality and multimodal fusion methods; (ii) validation of the proposed 3DPCS strategy by comparing it with a baseline point cloud projection method; and (iii) evaluation of the IB-based latent representation, focusing on its compression efficiency, task-relevant feature extraction, and robustness to sensor noise.

\subsection{Experiment Setup}
Experiments are conducted on the MMAUD dataset \cite{yuan2024MMAUD}, where stereo RGB images and LiDAR point clouds are paired by nearest timestamps and annotated with 3D UAV positions. The dataset is split into training and testing sets at a ratio of 8:2, with five-fold cross-validation on the training set.

The implementation is based on PyTorch and trained on a server with an Intel Xeon Platinum 8350C CPU and an NVIDIA GeForce RTX 4090 GPU. The Adam optimizer \cite{kingma2014adam} is used with a weight decay of $5 \times 10^{-4}$ and a batch size of 32. Early stopping is applied to alleviate overfitting. For multimodal training, a differential learning rate strategy is adopted: pretrained encoders are fine-tuned with a learning rate of $1 \times 10^{-5}$, while newly added modules are trained with $1 \times 10^{-4}$. Detailed network settings are listed in Table~\ref{tab:network_construction}.

\begin{table*}[t]
\vspace{0.05in}
\caption{Detailed Construction of the Proposed Network Modules.}
\centering
\label{tab:network_construction}
\renewcommand{\arraystretch}{1.1}
\begin{tabular}{|l|l|}
\hline
\textbf{Module} & \textbf{Architectural Construction and Mathematical Operation} \\ \hline
Stereo & $\text{ReLU}([\text{RN18}_{[:-1]}(\mathbf{I}^{\mathrm{L},\text{crop}}), \text{RN18}_{[:-1]}(\mathbf{I}^{\mathrm{R},\text{crop}}), \text{MLP}_{8\to 64 \to 128}(\mathbf{b})] \to \text{Linear}_{1152 \to 256} )$ \\ 
Encoder & $ \to \text{Drop}(0.3) \to \text{Linear}_{256 \to 128} \to \text{ReLU}$ \\ \hline
Depth Stem & $\text{ReLU}\left(\text{BN}(\text{Conv}_{5\to 16}^{3\times 3})\right)  \to \text{ReLU}\left(\text{BN}(\text{Conv}_{16\to 32}^{3\times 3})\right) \to \text{Conv}_{32\to 3}^{1\times 1}$ \\ \hline
Depth Encoder & $ \text{ReLU} \left( \text{Stem}_{5\to 3} \to \text{RN18}_{\text{conv1}\to\text{layer4}} \to \text{AvgPool}_{7 \to 1} \to \text{Linear}_{512 \to 128} \right) \to \text{Drop}(0.2)$ \\ \hline
Cross Fusion  & $\mathbf{w}^{(m)} = \sigma\left(\text{Linear}_{128 \to 128}\left(\text{ReLU}(\text{Linear}_{256 \to 128}([\mathbf{x}^{\mathrm{c}}, \mathbf{x}^{\mathrm{l}}]))\right)\right),\quad m \in \{\mathrm{c}, \mathrm{l}\}$ \\ \hline
Pre-branch & $\mathbf{x}^{\text{pre}} = \text{ReLU}\left(\text{Linear}_{256 \to 128}([\mathbf{x}^{\mathrm{c}} \odot (1 + \mathbf{w}^\mathrm{c}), \mathbf{x}^{\mathrm{l}} \odot (1 + \mathbf{w}^{\mathrm{l}})])\right)$ \\ \hline
Information Bottleneck & $\boldsymbol{\mu} = \text{Linear}_{128 \to k}(\mathbf{x}^{\text{pre}}), \quad \log\boldsymbol{\sigma}^2 = \text{Linear}_{128 \to k}(\mathbf{x}^{\text{pre}}) \quad \text{s.t.} \quad \mathbf{z}^\mathrm{n} = \boldsymbol{\mu} + \boldsymbol{\epsilon} \odot \exp(0.5 \cdot \log\boldsymbol{\sigma}^2), \quad \boldsymbol{\epsilon} \sim \mathcal{N}(\mathbf{0}, \mathbf{I})$ \\ \hline
Auxiliary Heads & $\hat{\mathbf{y}}^{\mathrm{l}} = \text{Linear}_{64 \to 3}(\text{ReLU}(\text{Linear}_{128 \to 64}(\mathbf{x}^{\mathrm{l}}))), \quad \hat{\mathbf{y}}^{\mathrm{c}} = \text{Linear}_{128 \to 3}(\mathbf{x}^{\mathrm{c}})$ \\ \hline
Task-specific Regressor & $\hat{\mathbf{y}} = \hat{\mathbf{y}}^{\mathrm{c}} + \text{Linear}_{64\to 3}\left(\text{ReLU}(\text{Linear}_{k \to 64}(\mathbf{z}^{\mathrm{n}}))\right) \quad (\text{Stereo Base Estimate} + \text{IB Residual Refinement})$ \\ \hline
\end{tabular}
\end{table*}

\begin{figure}[!t]
\centering
\vspace{-1ex}
\includegraphics[width=0.85\linewidth]{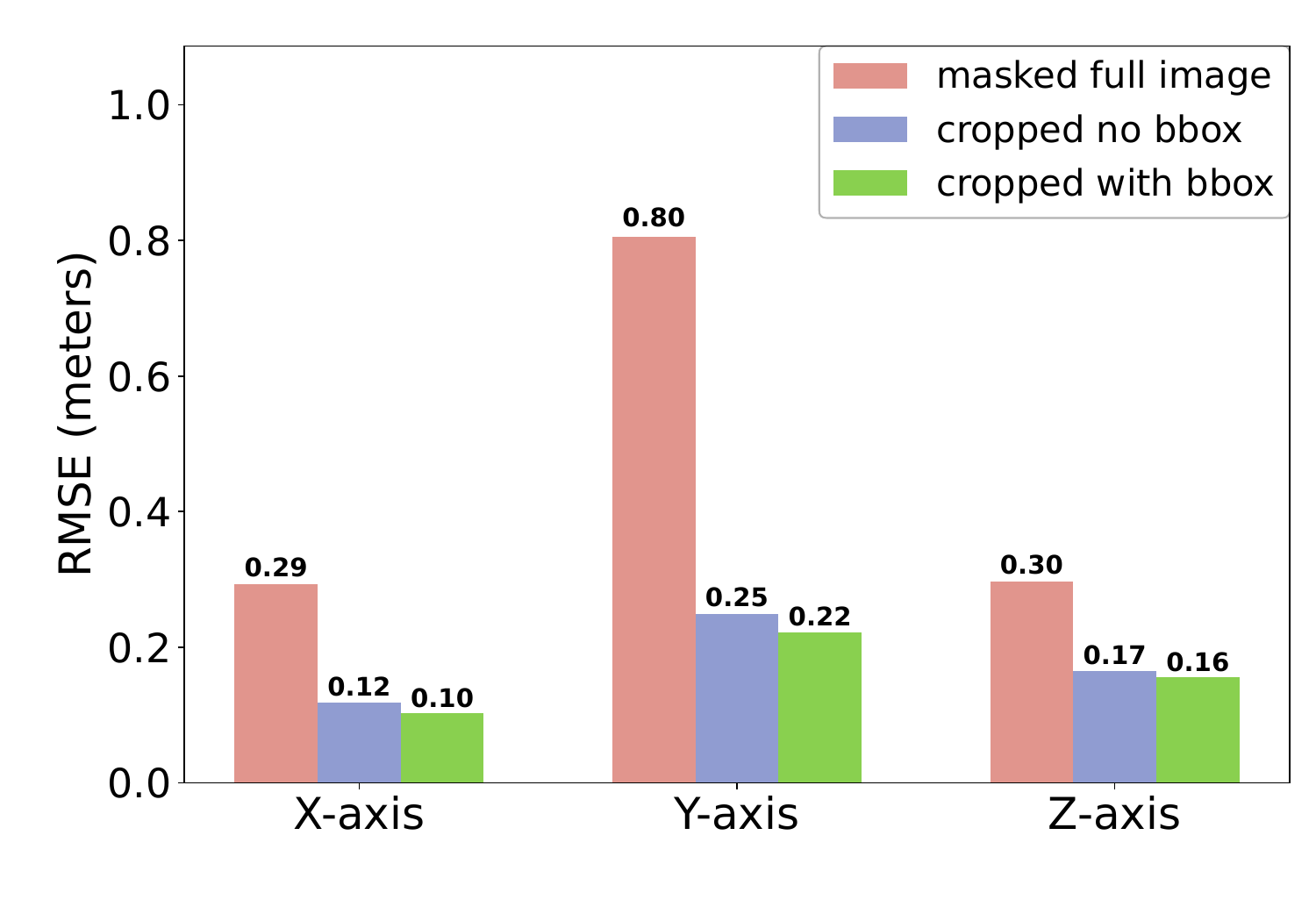}
\vspace{-2ex}
\caption{Comparison of test-set RMSE values across various pre-processing pipelines.}
\label{model_rmse}
\centering
\includegraphics[width=0.85\linewidth]{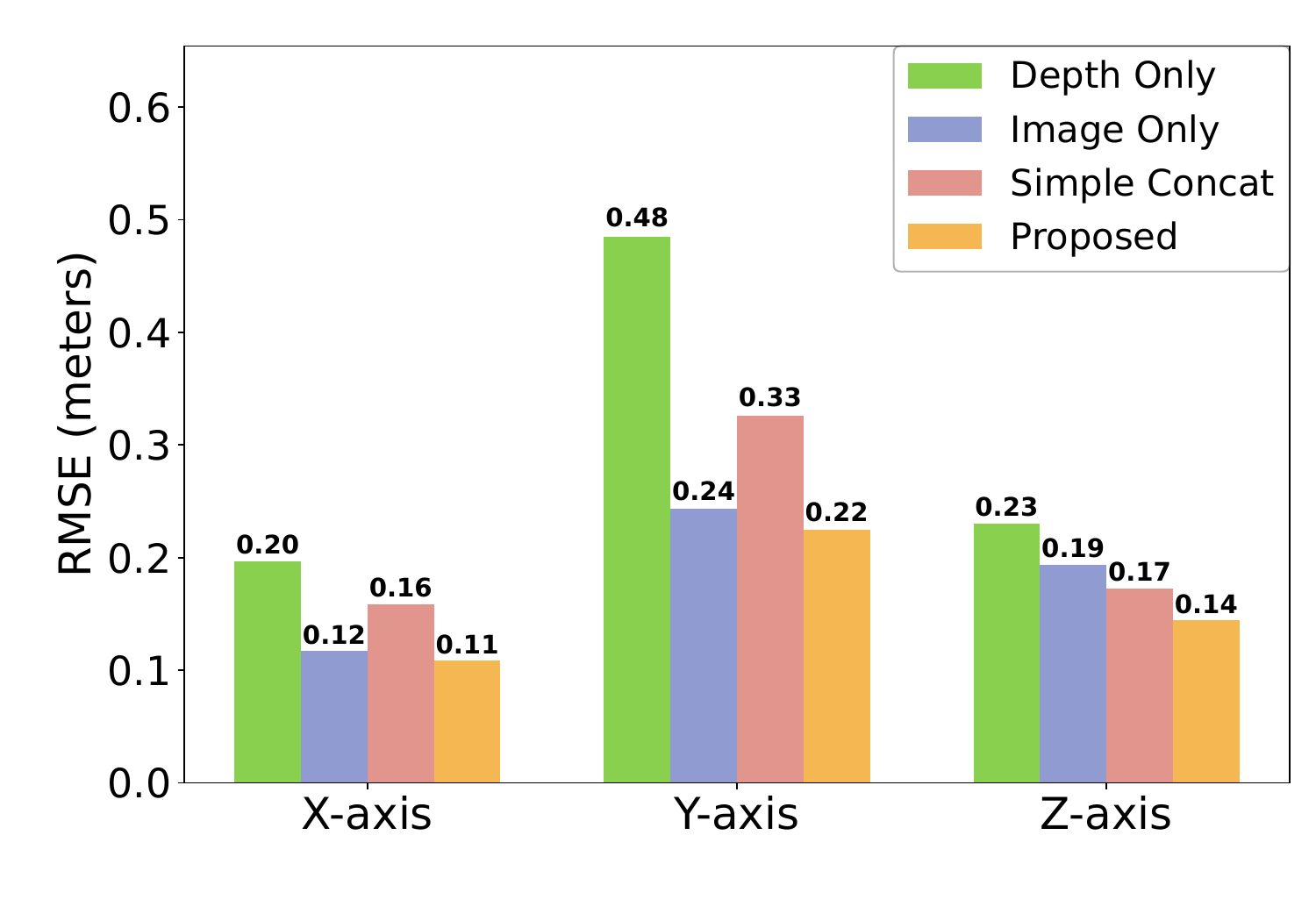}
\vspace{-3ex}
\caption{Comparison of test-set RMSE values across different methods.}
\label{prepocess_rmse}
\vspace{-2ex}
\end{figure}

\subsection{Data Processing and Fusion Strategy}

To validate the PRI-Net framework, we first compare different image input processing strategies under the single-modality setting, and then evaluate the performance of single-modality methods and the proposed multimodal fusion method.

For the image modality, three input strategies are compared: ROI-cropped images with bounding box information, ROI-cropped images only, and masked full-image input. Since the UAV occupies only a small portion of the full image, directly using the full image may introduce background interference. Therefore, masked full-image input is adopted for comparison. As shown in Fig.~\ref{model_rmse}, the ROI-cropped input with bounding box information achieves the best performance, while removing the bounding box information leads to slight degradation, and the masked full-image input yields the largest error. This shows that target-region cropping effectively suppresses background interference, while bounding box information further provides useful spatial priors for localization.

For the fusion strategy, Fig.~\ref{prepocess_rmse} compares the proposed RAF method with three baselines: Image Only and Depth Only, which use a single modality as input, and Simple Concat, which directly concatenates high-dimensional image and depth features for regression. The proposed method achieves the lowest root mean square error (RMSE) on all three axes. In particular, compared with Image Only, the Z-axis error is reduced by about 26.3\%, indicating that depth information effectively improves distance estimation. Meanwhile, Simple Concat even performs worse than Image Only on all three axes. This suggests that direct concatenation cannot fully exploit cross-modal complementarity, and modality imbalance may instead interfere with the dominant modality. These results further demonstrate the effectiveness of the proposed method.

\begin{figure}[!t]
\centering
\vspace{-1ex}
\includegraphics[width=0.85\linewidth]{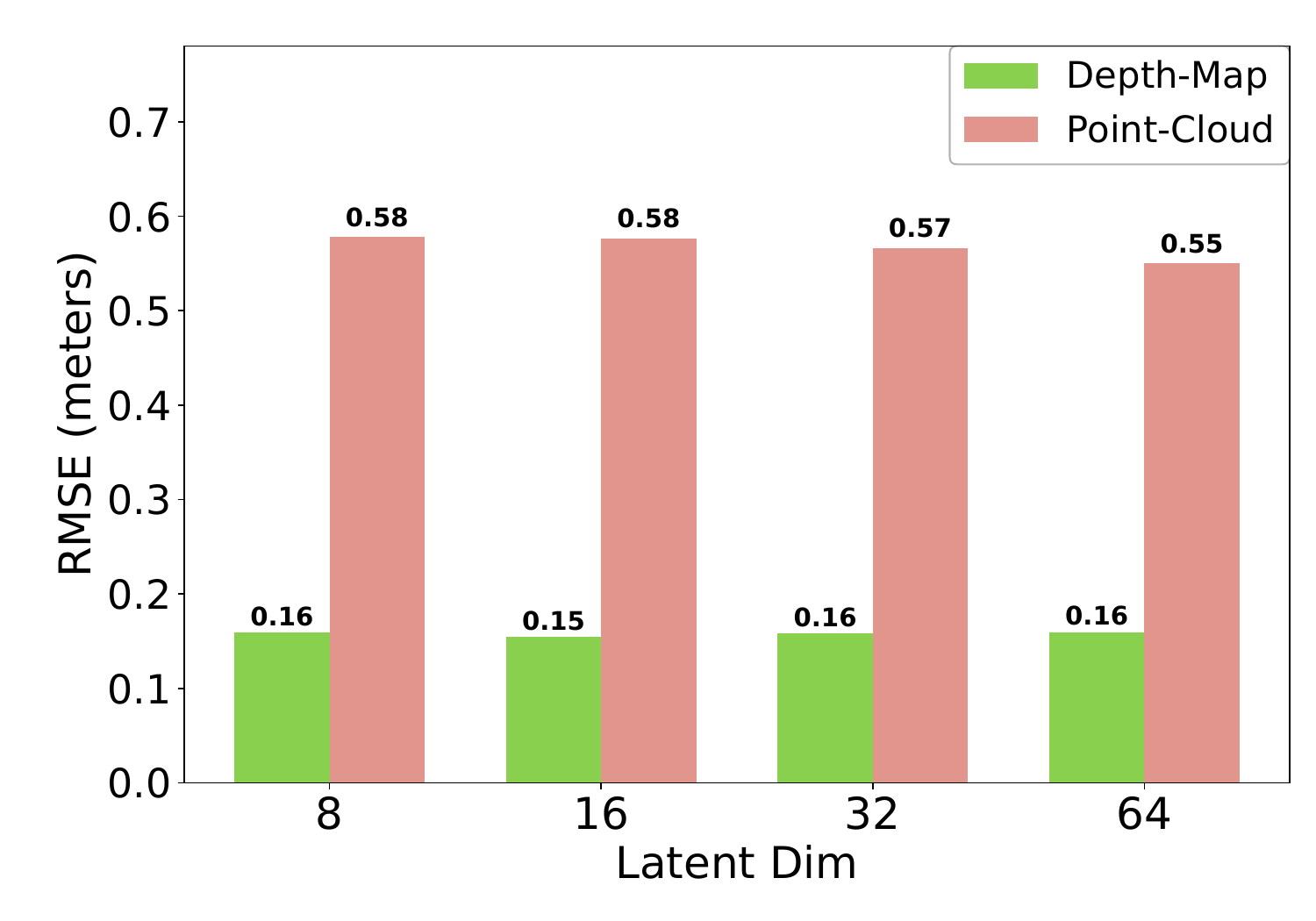}
\vspace{-2ex}
\caption{Comparison of test-set RMSE across different latent dimensionalities $k$.}
\label{fig_pc}
\centering
\includegraphics[width=0.85\linewidth]{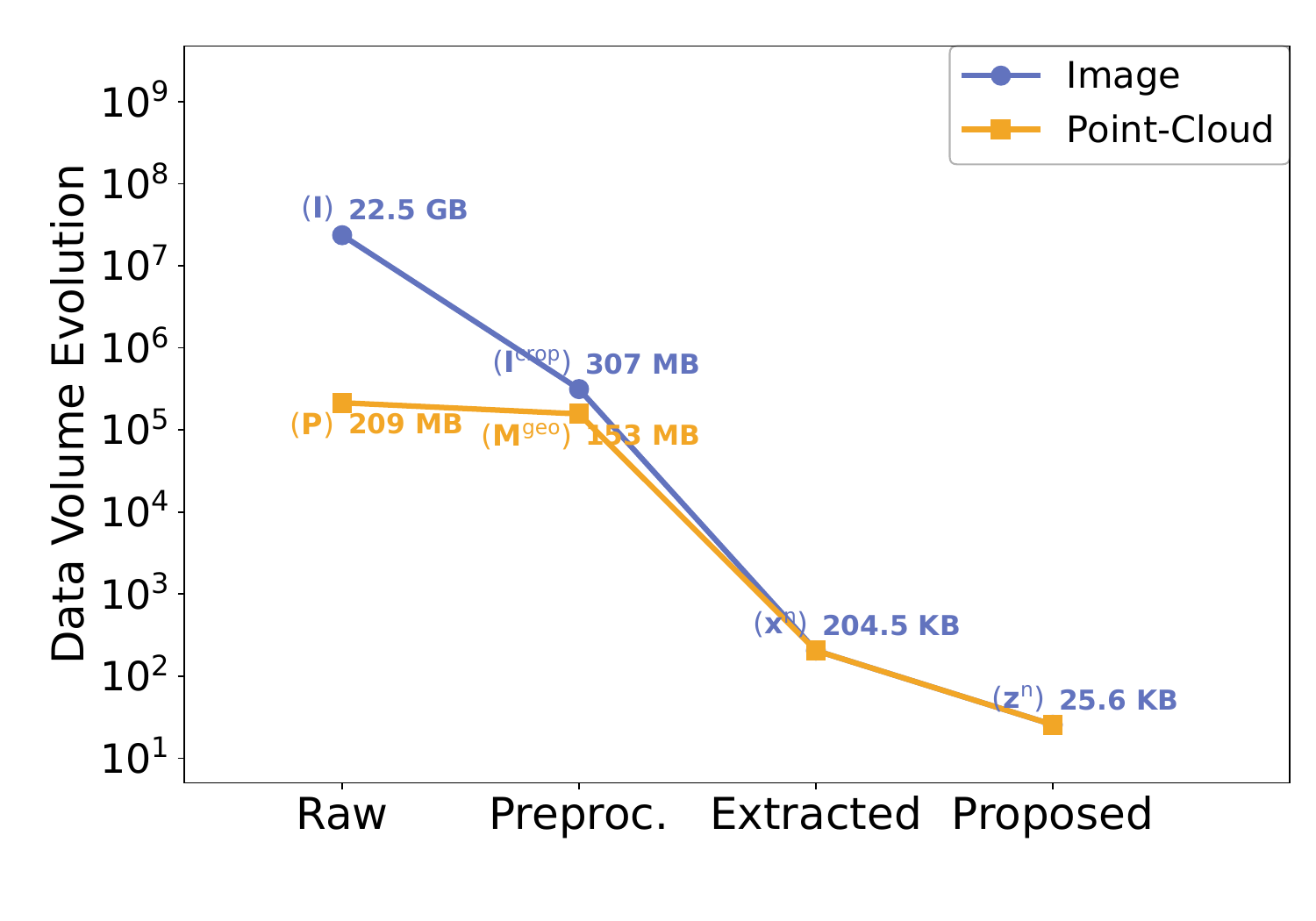}
\vspace{-3ex}
\caption{Evolution of data volume across different stages of the proposed multimodal localization framework.}
\label{fig_ib}
\vspace{-2ex}
\end{figure}

\begin{table}[htbp]
\vspace{0.05in}
\caption{Average Test RMSE Comparison}
\centering
\label{tab:rmse_comparison}
\footnotesize
\setlength{\tabcolsep}{2pt}
\begin{tabular*}{\columnwidth}{@{\extracolsep{\fill}}llcccc}
\toprule
\textbf{Noise Level} & \textbf{Model} & \multicolumn{4}{c}{\textbf{Latent Dimension ($d$)}} \\
\cmidrule(lr){3-6}
(m / $\sigma_{d}$) & \textbf{Type} & \textbf{8} & \textbf{16} & \textbf{32} & \textbf{64} \\
\midrule
None & IB & \textbf{0.1589} & \textbf{0.1542} & \textbf{0.1580} & 0.1595 \\
(Original) & MLP & 0.1629 & 0.1563 & 0.1596 & \textbf{0.1570} \\
\midrule
$\text{mask\_size}=20$ & IB & \textbf{0.2041} & \textbf{0.1820} & \textbf{0.1966} & 0.1908 \\
$\text{mask\_ratio}=0.1$ & MLP & 0.2133 & 0.1908 & 0.2014 & \textbf{0.1876} \\
\midrule
$\text{mask\_size}=20$ & IB & \textbf{0.2033} & \textbf{0.1867} & \textbf{0.2022} & 0.1923 \\
$\text{mask\_ratio}=0.2$ & MLP & 0.2062 & 0.1891 & 0.2089 & \textbf{0.1897} \\
\midrule
$\text{mask\_size}=40$ & IB & \textbf{0.2190} & \textbf{0.2084} & \textbf{0.2094} & \textbf{0.2113} \\
$\text{mask\_ratio}=0.1$ & MLP & 0.2204 & 0.2094 & 0.2310 & 0.2138 \\
\midrule
$\text{mask\_size}=40$ & IB & \textbf{0.2179} & \textbf{0.2010} & \textbf{0.2169} & \textbf{0.2081} \\
$\text{mask\_ratio}=0.2$ & MLP & 0.2231 & 0.2127 & 0.2302 & 0.2150 \\
\bottomrule
\end{tabular*}
\end{table}

\subsection{The Proposed 3D Point Cloud Splatting Strategy}

To validate the effectiveness of the 3DPCS strategy, we further compare the proposed depth-map-based pipeline with the point-cloud baseline under different latent dimensions. Specifically, we compare the RMSE of the two methods across various dimensions $k \in \{8, 16, 32, 64\}$. As illustrated in Fig.~\ref{fig_pc}, our method achieves a significantly lower and more stable error floor regardless of the latent dimension, whereas the baseline exhibits substantially higher errors and greater sensitivity to dimensionality changes. These gains come from 3DPCS as an offline geometric preprocessing step that adds no trainable parameters or inference layers, and from its structured grid representation for local continuity and discriminative spatial features.

\subsection{Latent Representation Efficiency}

Using a fixed $\beta$, we evaluate the effect of latent dimensionality $k$ and robustness to interference by comparing the IB-based scheme with an MLP regressor. Image interference is simulated by structural occlusion and depth interference by random signal loss. As shown in Table~\ref{tab:rmse_comparison}, IB achieves lower average RMSE when $k \le 32$, while the MLP baseline performs better at $k=64$ due to its stronger fitting capacity in higher-dimensional spaces. Under noise and partial occlusion, IB remains more robust in constrained latent spaces, whereas MLP is more sensitive to interference.

To evaluate efficiency for resource-constrained UAVs, Fig.~\ref{fig_ib} shows the single-modality data volume across processing stages. While the raw input is large (22.5 GB for images and 209 MB for point clouds), the EMIB-based latent representation $\mathbf{z}^\mathrm{n}$ ($k=32$) reduces it to only 25.55 KB per modality. This corresponds to a $4.4 \times 10^5$-fold reduction from the raw data and an additional 87.5\% reduction from the intermediate feature $\mathbf{x}^\mathrm{n}$ (204.5 KB per modality).

These results show that the proposal effectively removes task-irrelevant redundancy and preserves compact task-relevant features. By converting heterogeneous raw inputs into a balanced latent representation, the IB-based scheme enables robust and communication-efficient real-time edge deployment.

\section{Conclusion}
\label{conclusion}


In this paper, we propose PRI-Net framework integrating image semantics, depth geometry, and information bottleneck compression. Experiments show that target-region image input reduces background interference, while point cloud splatting improves depth representation and localization accuracy. Compared with unimodal baselines and conventional fusion methods, the proposed method achieves better performance on all axes, including a 26.3\% reduction in Z-axis error over the image-only model. In addition, the information bottleneck improves compression efficiency and noise robustness in low-dimensional representations, reducing transmission overhead while preserving localization accuracy. These results demonstrate the framework’s potential for resource-constrained edge deployment. Future work will consider robust sensing under dynamic scenes and occlusion, and adaptive $\beta$ tuning for compression under time-varying bandwidth.

\bibliographystyle{IEEEtran} 
\bibliography{refs}

\end{document}